\pdfoutput=1   % required by arXiv when the build must use pdflatex

\documentclass[11pt]{article}

\usepackage[letterpaper,left=1.15in,right=1.15in,top=1.0in,bottom=1.15in]{geometry}

\usepackage[T1]{fontenc}
\usepackage[utf8]{inputenc}
\usepackage{mathptmx}                    % math
\usepackage{charter}                     % body text (Bitstream Charter)
\usepackage[scaled=0.90]{helvet}         % headings / logo captions
\usepackage{microtype}

\usepackage{graphicx}
\usepackage[table,dvipsnames]{xcolor}
\usepackage[most]{tcolorbox}
\usepackage{tikz}

\usepackage{amsmath}
\usepackage{amsfonts}
\usepackage{amssymb}

\usepackage{booktabs}
\usepackage{multirow}
\usepackage{tabularx}
\usepackage{array}

\usepackage{titlesec}
\usepackage[labelfont=bf,labelsep=quad,font=small,skip=6pt]{caption}
\usepackage{enumitem}
\usepackage{float}
\usepackage{placeins}
\usepackage{xurl}

\usepackage[numbers,sort&compress]{natbib}

\definecolor{omblue}{RGB}{29,86,190}      % OpenMind blue  (headings/links)
\definecolor{omgreen}{RGB}{78,216,155}
\definecolor{omorange}{RGB}{255,165,31}
\definecolor{ompink}{RGB}{245,0,110}
\definecolor{omrule}{RGB}{200,204,210}
\definecolor{omgray}{RGB}{110,115,125}

\usepackage[colorlinks=true,
            linkcolor=omblue,
            citecolor=omblue,
            urlcolor=omblue,
            filecolor=omblue,
            breaklinks=true,
            pdfborder={0 0 0}]{hyperref}

\titleformat{\section}
  {\sffamily\bfseries\large\color{omblue}}{\thesection}{0.7em}{}
\titleformat{\subsection}
  {\sffamily\bfseries\normalsize\color{omblue}}{\thesubsection}{0.6em}{}
\titleformat{\subsubsection}
  {\sffamily\bfseries\small\color{omblue}}{\thesubsubsection}{0.6em}{}
\titlespacing*{\section}{0pt}{16pt plus 3pt minus 2pt}{7pt}
\titlespacing*{\subsection}{0pt}{12pt plus 2pt minus 2pt}{5pt}

\setlist{itemsep=2pt,topsep=4pt,parsep=0pt,leftmargin=1.4em}

\newcommand{\brandrule}{%
  \noindent\begin{tikzpicture}
    \fill[ompink]   (0,0)                rectangle (0.25\linewidth,1.1pt);
    \fill[omorange] (0.25\linewidth,0)   rectangle (0.50\linewidth,1.1pt);
    \fill[omblue]   (0.50\linewidth,0)   rectangle (0.75\linewidth,1.1pt);
    \fill[omgreen]  (0.75\linewidth,0)   rectangle (\linewidth,1.1pt);
  \end{tikzpicture}}

\newcommand{\missingfig}[2]{%
  {\setlength{\fboxsep}{0pt}\setlength{\fboxrule}{0.6pt}%
   \textcolor{omrule}{\fbox{%
     \begin{minipage}[c][#2][c]{\dimexpr\linewidth-1.2pt\relax}
       \centering\sffamily\small\color{omgray}
       figure file not found\\[3pt]{\ttfamily\small\detokenize{#1}}
     \end{minipage}}}}}

\newcommand{\incfig}[3][\linewidth]{% [width]{path}{placeholder height}
  \IfFileExists{#2}{\includegraphics[width=#1]{#2}}{\missingfig{#2}{#3}}}

\newcommand{\PaperTitle}{SocioGesture: Real-Time and Adaptive Social Gesture\\
Perception for Human-Robot Interaction}
\newcommand{\ProjectPage}{https://wenjinfu.github.io/socioGesture/}
\newcommand{\PaperDate}{\today}

\hypersetup{
  pdftitle={SocioGesture: Real-Time and Adaptive Social Gesture Perception for Human-Robot Interaction},
  pdfauthor={Wenjin Fu and Li-Fan Wu and Jerin Peter and Chip Huyen and Boyuan Chen and Jan Liphardt},
  pdfsubject={Human-Robot Interaction, Gesture Recognition, Robot Perception},
  pdfkeywords={human-robot interaction, gesture recognition, skeleton action recognition, edge deployment}
}

\begin{document}

% =====================================================================
%  HEADER: OpenMind logo + brand rule
% =====================================================================
\thispagestyle{empty}

\noindent\includegraphics[height=19pt]{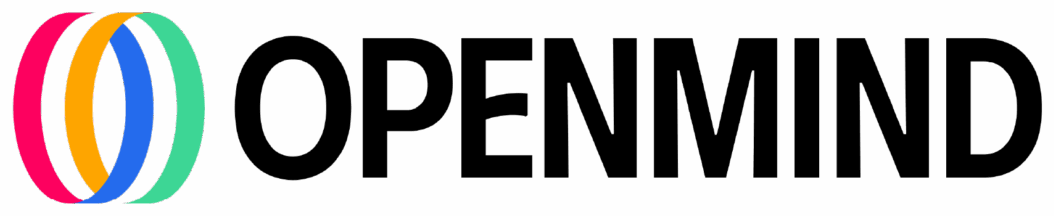}\hfill\mbox{}

\vspace{6pt}
\brandrule
\vspace{18pt}

% =====================================================================
%  TITLE BLOCK  --  REPLACE THE AUTHOR NAMES BELOW
% =====================================================================
\begin{center}
  {\fontsize{17.5}{22}\selectfont\bfseries \PaperTitle\par}

  \vspace{16pt}

  % ---------------- AUTHORS -----------------------------------------
  {\large
   \textbf{Wenjin Fu},\;
   \textbf{Li-Fan Wu},\;
   \textbf{Jerin Peter},\;
   \textbf{Chip Huyen},\;
   \textbf{Boyuan Chen},\;
   \textbf{Jan Liphardt}\par}

  \vspace{9pt}

  % ---------------- AFFILIATION -------------------------------------
  {\normalsize OpenMind\par}

  \vspace{5pt}

  {\small\color{omgray}
   \href{mailto:wendy@openmind.com}{\texttt{\{wendy, lifan, jerin, chip, boyuan, jan\}@openmind.com}}\par}
\end{center}

\vspace{10pt}

% =====================================================================
%  ABSTRACT BOX
% =====================================================================
\begin{tcolorbox}[
    enhanced, breakable,
    colback=white, colframe=omblue!55!white,
    boxrule=0.7pt, arc=6pt,
    left=14pt, right=14pt, top=10pt, bottom=10pt]

  \begin{center}
    {\sffamily\bfseries\large\color{omblue} Abstract}
  \end{center}
  \vspace{-2pt}

  \small
  Robots interacting with people must recognize not only explicit commands, but also
  social cues such as invitations, refusals, and unavailability. In real deployments,
  these cues must be inferred from noisy onboard perception under partial occlusion,
  changing viewpoints, and strict latency constraints. We present \textbf{SocioGesture},
  a real-time adaptive social gesture perception system for human-robot interaction (HRI).
  SocioGesture uses a compact confidence-aware body-hand skeleton representation and a
  lightweight dual-stream model that fuses body motion with hand articulation for
  low-latency onboard recognition. To improve deployment robustness, we train the model
  with occlusion-aware skeleton corruption, exposing it to missing hands, occluded arms,
  and temporally unstable keypoints without increasing the inference cost. On a social
  gesture dataset collected in mixed indoor--outdoor HRI scenarios, SocioGesture achieves
  strong held-out-subject recognition, substantially improves robustness under structured
  joint occlusion, and runs in real time on a robot-mounted edge device. During deployment,
  uncertain interaction segments are saved for offline labeling and adaptation, enabling
  SocioGesture to expand its gesture vocabulary while preserving performance in the
  original classes. These results demonstrate a practical path toward robust, efficient,
  and adaptive social perception for interactive robots.

  \vspace{6pt}

  \vspace{4pt}
  \textbf{Date:} \PaperDate\\
  \textbf{Project Page:} \url{\ProjectPage}
\end{tcolorbox}

\vspace{6pt}

% =====================================================================
\section{Introduction}
% =====================================================================
Human-robot interaction (HRI) often begins before any explicit verbal command. A person may wave to attract attention, beckon a robot closer, raise a hand to stop it, or signal unavailability through everyday gestures. Misinterpreting such cues can cause socially inappropriate or unsafe behavior: a false invitation may trigger unwanted approach, while a missed stop signal may fail to halt the robot. We focus on a greeter-style HRI setting where the robot should engage available people while deferring under prohibition, unavailability, or absence of attention. Unlike offline action classification, gesture recognition for HRI directly affects robot behavior and must be reliable under onboard perception noise.

Recognizing social gestures on a robot differs from curated video recognition. The robot observes people from moving or constrained viewpoints, with varying distance, lighting, self-occlusion, and intermittent keypoint failures. Hands and arms are especially important because they carry many social cues but are small and often occluded. The system must also be low latency: delayed responses disrupt interaction flow and make the robot appear unresponsive. Finally, deployment interactions are open-ended; a practical system should improve when it encounters uncertain or previously unseen gesture styles. A key difficulty is that these requirements conflict: models large enough to reason over rich video context may be too slow or cloud-dependent for safe interaction, while small onboard models may be brittle when the pose estimator fails.

Skeleton-based action recognition provides a useful foundation because keypoints reduce sensitivity to clothing, background, and illumination. However, prior skeleton-based action recognition methods are typically evaluated offline on benchmark datasets, where recognition is decoupled from robot control~\cite{stgcn,pyskl,ctrgcn,infogcn,blockgcn}. HRI instead requires a deployable perception-control loop that is robust to missing joints, efficient enough for closed-loop behavior, and capable of post-deployment improvement. SocioGesture resolves this tension by separating immediate behavior selection from slower deployment adaptation: the robot acts only on local, high-confidence skeleton predictions and stores uncertain interaction segments for later analysis instead of forcing potentially unsafe decisions.

\begin{figure}[t]
    \centering
    \incfig{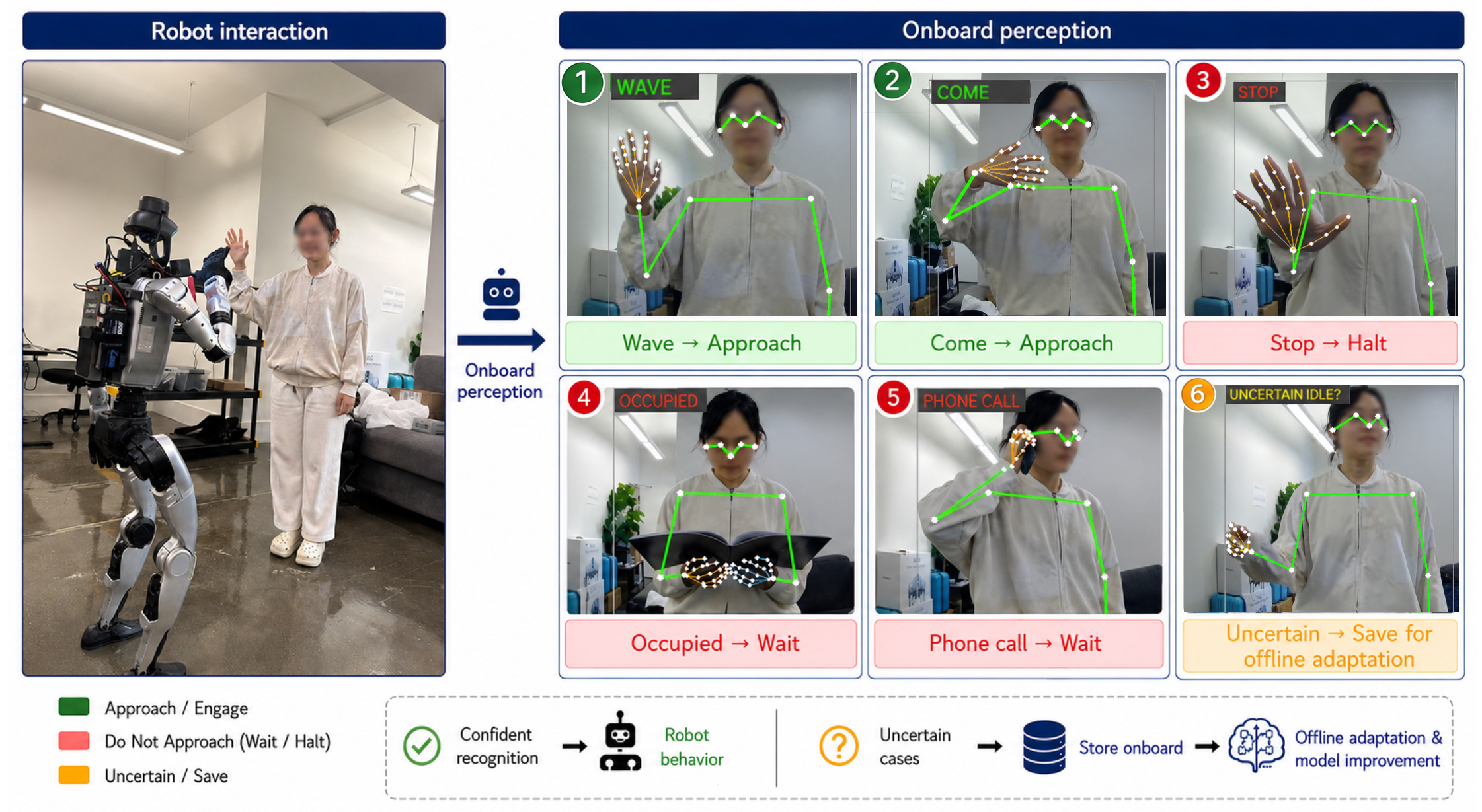}{52mm}
    \caption{\textbf{SocioGesture enables real-time and adaptive social gesture perception for human-robot interaction.} Confident predictions are mapped to conservative robot behaviors, while uncertain interaction segments are saved for offline labeling and model adaptation.}
    \label{fig:teaser}
\end{figure}

We present SocioGesture, a real-time and adaptive social gesture perception system for HRI (Figure~\ref{fig:teaser}). SocioGesture keeps closed-loop recognition lightweight and local, while using stronger video understanding models only offline to label uncertain interaction segments for adaptation.

Our contributions are as follows:

\begin{itemize}
  \item \textbf{A deployable social gesture perception system} that connects onboard body-hand pose estimation to conservative greeter-robot behaviors in a closed loop.
  \item \textbf{A compact confidence-aware skeleton representation} and a lightweight body-hand fusion model evaluated in a single forward pass for low-latency edge inference.
  \item \textbf{Occlusion-aware training} that improves robustness to missing hands and arms without any inference-time overhead.
  \item \textbf{An uncertainty-driven offline adaptation loop} that expands the gesture vocabulary while preserving original-class performance.
\end{itemize}

Our dataset and evaluation code will be released at
\url{https://wenjinfu.github.io/socioGesture/} to support future research
on deployable HRI perception.

% =====================================================================
\section{Related Work}
% =====================================================================

\subsection{Skeleton-Based Gesture and Action Recognition}
Skeleton-based action recognition represents human activity as temporal sequences of body joints, reducing sensitivity to appearance, clothing, illumination, and background variation. Spatial-temporal graph convolutional networks established graph-based skeleton modeling as a standard framework~\cite{stgcn}, and later methods improve training practice, graph topology, representation learning, and topology awareness~\cite{pyskl,ctrgcn,infogcn,blockgcn}. Heatmap-based and body-hand methods further improve recognition by changing the pose representation or modeling hand articulation~\cite{poseconv3d,bharnet}; the body-hand cross-attention design of~\cite{bharnet} is closest to ours, but it fuses multiple inference passes and, like the methods above, is evaluated on curated benchmarks decoupled from robot behavior, latency, and deployment-time perception failures. SocioGesture instead targets robot-view social gestures with a compact confidence-aware body-hand model evaluated in a single forward pass for low-latency edge inference.

\subsection{Robust and Deployable Gesture Perception for HRI}
Gestures have long served as a non-verbal communication channel in human-robot interaction, supporting command following, teleoperation, collaboration, and navigation~\cite{waldherr2000gesture,wang2022handarm}. Gesture-aware navigation and multi-robot interaction systems show how pointing, beckoning, and related body cues can be mapped to robot motion or navigation policies~\cite{alonsomora2015gesture}, while online recognition work highlights the need for timely predictions from partial observations~\cite{saponaro2013anticipation}. In deployment, robot-mounted pose estimates can be incomplete or unstable due to self-occlusion, viewpoint changes, distance, and pose-estimation failures. Prior skeleton-recognition work improves robustness through pose heatmaps, graph architectures, confidence cues, and training-time perturbations~\cite{poseconv3d,pyskl,ctrgcn,bharnet}, but usually studies these techniques offline rather than in a closed-loop robot perception system. SocioGesture addresses this deployment setting by training with occlusion-aware skeleton corruption, improving robustness to missing hands, occluded arms, and unstable keypoints without adding inference-time computation.

\subsection{Foundation Models and Deployment Adaptation}
Foundation models have been used in robotics for semantic perception, planning, language-conditioned control, and data generation~\cite{saycan,palme,rt2}. Vision-language and video understanding models provide rich semantic interpretation, but using them as the primary perception module for time-critical HRI can introduce latency, compute cost, and dependence~\cite{zheng2025diffusion}. Active and continual learning offer another route for improving robot perception after deployment through uncertainty-driven sample selection and incremental adaptation~\cite{active_learning_robotics,continual_learning_autonomous,zheng2025cafll}. SocioGesture combines these ideas in an online--offline design: a lightweight skeleton model runs onboard, while uncertain interaction segments are labeled offline by a stronger video model and used to fine-tune the recognizer.

% =====================================================================
\section{Method}
\label{sec:method}
\begin{figure}[t]
    \centering
    \incfig{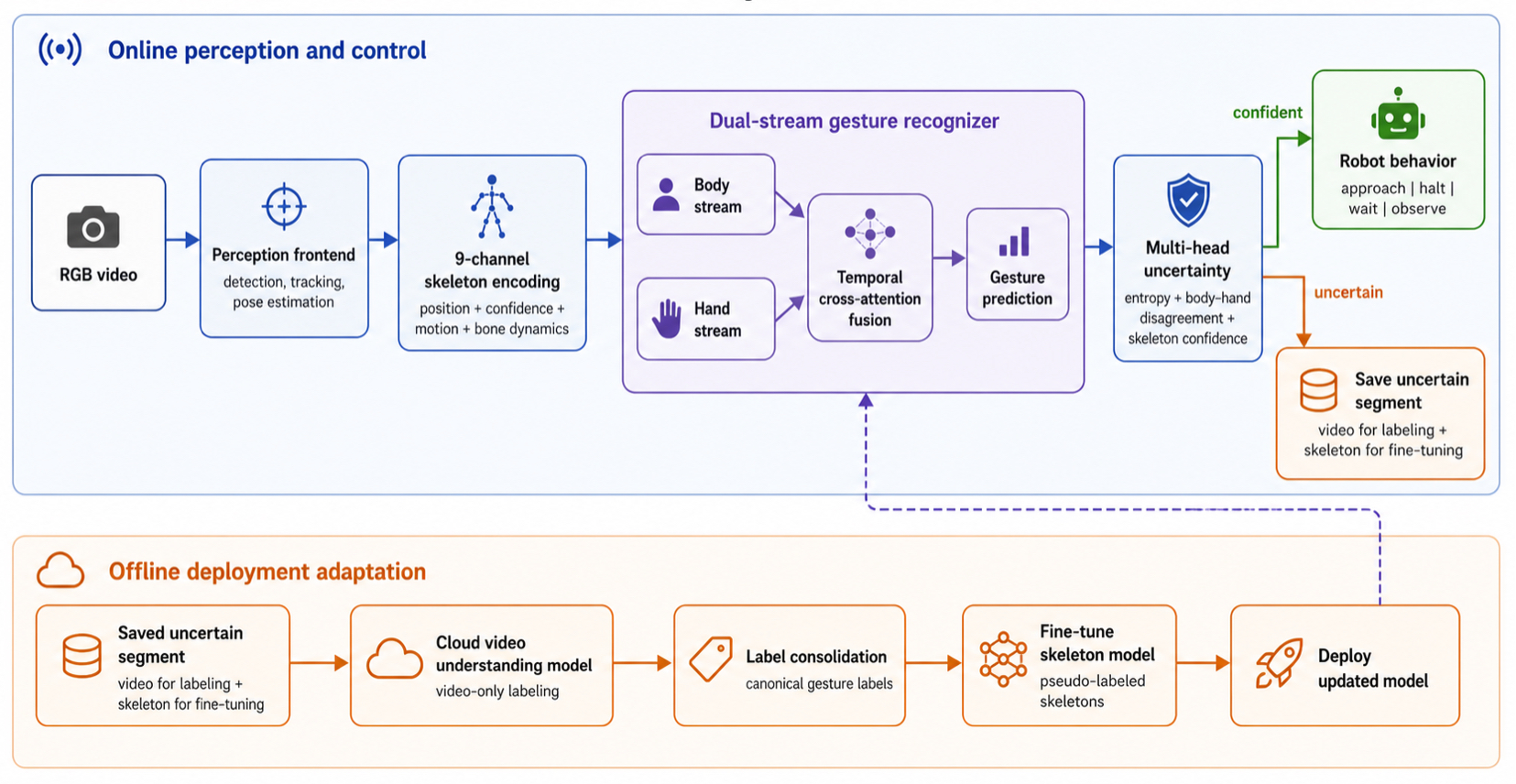}{58mm}
    \caption{\textbf{System overview.} The online path converts RGB video into confidence-aware skeleton windows and predicts gestures with a dual-stream body-hand model. Confident predictions are mapped to conservative robot behaviors, while uncertain interaction segments are saved with paired RGB video and skeleton data. The offline path labels saved videos, consolidates labels, fine-tunes the skeleton model, and redeploys the updated recognizer.}
    \label{fig:system}
\end{figure}

% =====================================================================
SocioGesture has an online path for real-time interaction and an offline path for adaptation (Figure~\ref{fig:system}). During online operation, the RGB stream is processed by person detection, tracking, and pose estimation. For each tracked person, a sliding skeleton window is encoded and classified by a lightweight gesture model: high-confidence predictions are mapped to conservative robot behaviors, while low-confidence segments are saved for later labeling and fine-tuning. The online and offline paths share the same skeleton representation, allowing interaction segments collected during deployment to be used directly for offline adaptation.

\subsection{Confidence-Aware Skeleton Encoding}
For each frame $t$ and joint $j$, the pose estimator provides a 2D coordinate $\mathbf{p}_{t,j}\in\mathbb{R}^{2}$ and a confidence score $c_{t,j}\in[0,1]$. Rather than discarding windows with missing or unreliable joints, SocioGesture retains each joint's coordinate together with its confidence, so partially observed windows remain usable. Given a window of $T$ frames and $J$ joints, SocioGesture constructs a 9-channel joint feature:
\begin{equation}
\label{eq:skeleton_feature}
    \mathbf{x}_{t,j}=\left[\mathbf{p}_{t,j},\, c_{t,j},\, \mathbf{p}_{t,j}-\mathbf{p}_{t-1,j},\, \mathbf{b}_{t,j},\, \Delta\mathbf{b}_{t,j}\right]\in\mathbb{R}^{9},
\end{equation}
where $\pi(j)$ denotes the parent of joint $j$ in the skeleton tree, $\mathbf{b}_{t,j}=\mathbf{p}_{t,j}-\mathbf{p}_{t,\pi(j)}$ is the corresponding bone vector, and $\Delta\mathbf{b}_{t,j}=\mathbf{b}_{t,j}-\mathbf{b}_{t-1,j}$ is the bone motion. Motion and bone-motion terms are zero at the first frame; for root joints without a parent, bone terms are zero. Coordinates are normalized within each tracked-person window before feature construction, making recognition depend primarily on relative posture and motion. After training-time corruption, motion, bone, and bone-motion channels are recomputed. The confidence channel helps distinguish an unreliable or missing hand from a stationary visible hand under partial occlusion.

\subsection{Body-Hand Gesture Model and Uncertainty}
Social gestures combine coarse body posture with fine hand articulation. SocioGesture therefore separates body and hand inputs, $\mathbf{X}^{B}\in\mathbb{R}^{T\times J_B\times 9}$ and $\mathbf{X}^{H}\in\mathbb{R}^{T\times J_H\times 9}$, and encodes them with lightweight graph-temporal backbones $f_B$ and $f_H$:
\begin{equation}
    \mathbf{F}^{B}=f_B(\mathbf{X}^{B}),\qquad \mathbf{F}^{H}=f_H(\mathbf{X}^{H}),
\end{equation}
where $\mathbf{F}^{B},\mathbf{F}^{H}\in\mathbb{R}^{T\times d}$ are temporally aligned body and hand features after spatial aggregation and projection. The two streams exchange information through temporal cross-attention,
\begin{equation}
    \widetilde{\mathbf{F}}^{B}=\mathrm{Attn}(\mathbf{F}^{B},\mathbf{F}^{H},\mathbf{F}^{H}),\qquad
    \widetilde{\mathbf{F}}^{H}=\mathrm{Attn}(\mathbf{F}^{H},\mathbf{F}^{B},\mathbf{F}^{B}),
\end{equation}
where $\mathrm{Attn}(Q,K,V)$ denotes multi-head attention over the temporal dimension. The attended features are temporally average-pooled into a fused representation $\mathbf{z}\in\mathbb{R}^{2d}$ and classified by a fused head. Auxiliary body and hand heads provide additional supervision during training and yield per-stream predictions used by the uncertainty score in Eq.~\eqref{eq:uncertainty}. Although the recognizer contains separate body and hand branches, it is evaluated as a single end-to-end forward pass, keeping inference cost low for edge deployment. This differs from multi-stream skeleton ensembles that run separate joint, bone, and motion models and fuse their predictions at inference time.

For behavior selection, the robot uses the fused maximum softmax probability. When confidence exceeds a fixed deployment threshold, gestures are mapped to conservative behaviors: invitation cues trigger approach, stop halts motion, and unavailability cues cause the robot to wait or maintain distance. In parallel, a deployment uncertainty score selects informative segments for offline labeling:
\begin{equation}
\label{eq:uncertainty}
    u = \alpha\,\frac{H(\mathbf{p})}{\log K} + \beta\, \tilde{D}(\mathbf{p}^{B},\mathbf{p}^{H}) + \gamma\,(1-\bar{c}),
\end{equation}
where $\mathbf{p}=\mathrm{softmax}(\mathbf{o})$ is the fused prediction and $H(\mathbf{p})/\log K$ is the entropy normalized by the number of gesture classes $K$. The disagreement term is $\tilde{D}(\mathbf{p}^{B},\mathbf{p}^{H})=D_{\mathrm{sym}}/(1+D_{\mathrm{sym}})$, where $D_{\mathrm{sym}}=\frac{1}{2}\left[\mathrm{KL}(\mathbf{p}^{B}\Vert\mathbf{p}^{H})+\mathrm{KL}(\mathbf{p}^{H}\Vert\mathbf{p}^{B})\right]$ and $\mathbf{p}^{B},\mathbf{p}^{H}$ are the auxiliary body and hand predictions. Here $\bar{c}$ is the average skeleton confidence in the window, and $\alpha,\beta,\gamma\!\ge\!0$ with $\alpha+\beta+\gamma=1$ weight the entropy, disagreement, and confidence terms. Specific hyperparameter weights ($\alpha{=}0.5$, $\beta{=}0.3$, $\gamma{=}0.2$) are reported in Appendix Table~\ref{tab:hyperparams}. This score is used only for uncertainty-driven segment selection, never as a behavior command; the behavior threshold is set conservatively for safety, while the selection score surfaces informative cases such as ambiguous hand-to-head motions, partial arm visibility, or body--hand disagreement.

\subsection{Occlusion-Aware Training and Offline Adaptation}
Robot-mounted pose estimates often fail when hands are self-occluded, arms leave the camera view, or keypoints become unstable. We train with paired clean and corrupted skeleton windows. Weak augmentation includes flipping, scaling, temporal shifting, and confidence-proportional coordinate noise. Structured corruption includes hand masking, arm-hand masking, temporal hand disappearance, and random joint masking. For corrupted joints, confidence is set to zero and coordinates are replaced with Gaussian noise; all motion, bone, and bone-motion channels are then recomputed.

For a view with fused logits $\mathbf{o}$, auxiliary logits $\mathbf{o}^{B},\mathbf{o}^{H}$, and label $y$, the supervised loss is
\begin{equation}
\label{eq:supervised_loss}
    \mathcal{L}_{\mathrm{sup}}=\mathcal{L}_{\mathrm{CE}}(\mathbf{o},y)+\lambda_{\mathrm{aux}}\left[\mathcal{L}_{\mathrm{CE}}(\mathbf{o}^{B},y)+\mathcal{L}_{\mathrm{CE}}(\mathbf{o}^{H},y)\right].
\end{equation}
Applying $\mathcal{L}_{\mathrm{sup}}$ to the clean and corrupted views yields $\mathcal{L}_{\mathrm{clean}}$ and $\mathcal{L}_{\mathrm{occ}}$. We also consider an optional consistency term that matches the corrupted-view prediction to the clean-view prediction:
\begin{equation}
\label{eq:consistency}
    \mathcal{L}_{\mathrm{cons}}
    = \tau^2 \mathrm{KL}\!\left(\mathrm{sg}(\mathbf{p}_{\mathrm{clean}}^{\tau})\,\|\,\mathbf{p}_{\mathrm{occ}}^{\tau}\right),
\end{equation}
where $\mathbf{p}^{\tau}=\mathrm{softmax}(\mathbf{o}/\tau)$, $\tau$ is the temperature, and $\mathrm{sg}(\cdot)$ stops gradients through the clean prediction. The full objective is
\begin{equation}
\label{eq:full_loss}
    \mathcal{L}=\mathcal{L}_{\mathrm{clean}}+\alpha_{\mathrm{occ}}\mathcal{L}_{\mathrm{occ}}+\rho(e)\,\alpha_{\mathrm{cons}}\mathcal{L}_{\mathrm{cons}},\qquad
    \rho(e)=\min\!\left(1,\frac{e}{E_w}\right),
\end{equation}
where $e$ is the current epoch and $\rho(e)$ ramps the consistency term over the first $E_w$ epochs to avoid penalizing unstable early predictions. Occlusion-aware training is applied only during training and adds no inference cost.

For adaptation, uncertain deployment segments $\mathcal{D}_{\mathrm{unc}}=\{(\mathbf{V}_n,\mathbf{X}_n)\}_{n=1}^{N}$ are labeled offline by a cloud video model, with $\mathbf{V}_n$ the saved RGB segment and $\mathbf{X}_n$ the corresponding skeleton sequence. After label consolidation and ambiguity filtering, the classifier is expanded from $K$ to $K_{\mathrm{new}}$ classes by copying original-class weights and initializing new-class weights. The model is fine-tuned on a mixture of original human-labeled windows and pseudo-labeled deployment windows, using the original data as rehearsal, and redeployed only after held-out validation.

% =====================================================================
\section{Experimental Evaluation}
% =====================================================================
We evaluate SocioGesture on five axes: held-out-subject recognition, robustness to structured occlusion, edge latency, offline vocabulary expansion, and live closed-loop deployment on the robot.

\subsection{Setup}
We collect a 7-class social gesture dataset in mixed indoor--outdoor HRI scenarios. The gesture set covers invitation (\textit{come}, \textit{wave}), prohibition (\textit{stop}), unavailability (\textit{phone call}, \textit{occupied}), and non-interaction (\textit{idle}, \textit{scratch head}). We include \textit{scratch head} as a distractor for hand-to-head gestures such as \textit{phone call}. These classes were selected because they map to distinct robot behavior modes rather than only visual categories: some invite engagement, some indicate availability, some block motion, and some signal that the person should not be interrupted. The dataset contains 10 subjects and 7055 skeleton windows. We use leave-one-subject-out (LOSO) cross-validation to measure generalization to unseen users. Skeleton sequences are segmented into 16-frame windows with 50\% overlap. Unless otherwise stated, HRI experiments use the 9-channel body-hand representation with a single-pass body-hand recognizer based on ST-GCN++~\cite{pyskl}. Each person is represented by $J=53$ joints: 11 upper-body joints and 42 hand joints. We use upper-body and hand joints because close-range social cues in our setting are primarily expressed through head, arm, and hand motion.

For deployment, the online perception stack runs on a humanoid robot with a robot-mounted NVIDIA Jetson Thor. Person detection uses YOLO11n~\cite{ultralytics_yolo11} exported to TensorRT FP16, tracking uses BoT-SORT~\cite{botsort}, and pose estimation uses RTMW-l~\cite{rtmw} exported to TensorRT FP16. We report full-stack latency rather than classifier-only latency because robot responsiveness depends on detection, tracking, pose estimation, skeleton construction, and gesture classification together. For adaptation, we evaluate on a held-out 10-class test set from five subjects not used for pseudo-label fine-tuning; these five subjects are disjoint from the 10 subjects used for LOSO recognition, and the five live-deployment participants (Section~\ref{sec:live}) are in turn disjoint from both. The expanded vocabulary adds \textit{thumbs up}, \textit{handshake}, and \textit{salute}.

\subsection{Recognition and Occlusion Robustness}
Table~\ref{tab:hri_occlusion} reports clean LOSO recognition and structured occlusion results on the HRI dataset. The standard model achieves $96.6\%\pm2.8\%$ clean accuracy, while occlusion-aware variants remain within one percentage point with overlapping fold-level variability, indicating comparable clean performance. Under structured occlusion, however, the difference is substantial: hand-occlusion accuracy improves from $31.8\%$ to $84.9\%$, and arm-occlusion accuracy improves from $28.3\%$ to $79.5\%$. The consistency variant gives the strongest arm and temporal-hand robustness, while the simpler occlusion-augmentation model provides nearly the same robustness with a simpler objective and is therefore used for adaptation. A controlled NTU RGB+D 60 stress test (Appendix~\ref{app:additional}) supports this interpretation: clean accuracy is preserved under occlusion-aware training, suggesting that the small HRI clean-accuracy differences mainly reflect LOSO fold variability rather than an inherent cost of the method.

\begin{table}[t]
\centering
\small
\setlength{\tabcolsep}{7pt}
\caption{\textbf{Clean recognition and structured occlusion evaluation on the HRI dataset.} The Clean column reports LOSO accuracy as mean $\pm$ standard deviation across 10 held-out-subject folds. Occlusion columns report mean $\pm$ standard deviation over three evaluation passes with independent occlusion realizations while keeping the trained models and test set fixed. Hand masks both hands, Arm masks one arm and its connected hand, and Temporal-Hand masks hand joints over a contiguous temporal span. All values are accuracies in percent.}
\label{tab:hri_occlusion}
\begin{tabular}{l c c c c}
\toprule
Training & Clean & Hand & Arm & Temporal-Hand \\
\midrule
Standard & $\mathbf{96.6 \pm 2.8}$ & $31.8 \pm 0.0$ & $28.3 \pm 0.2$ & $90.5 \pm 0.1$ \\
Occlusion Aug. & $95.7 \pm 3.2$ & $\mathbf{84.9 \pm 0.1}$ & $79.5 \pm 0.3$ & $92.6 \pm 0.1$ \\
Occlusion Consistency & $95.9 \pm 3.2$ & $84.6 \pm 0.1$ & $\mathbf{80.2 \pm 0.3}$ & $\mathbf{92.7 \pm 0.1}$ \\
\bottomrule
\end{tabular}
\end{table}

\subsection{Edge Latency and Offline Adaptation}
Table~\ref{tab:latency} reports full-stack deployment latency on the robot's onboard compute. The one-person stack runs at 25\,ms per frame without online cloud inference, within the 33\,ms budget of the 30\,FPS camera input; the three-person case runs at 42\,ms. In multi-person scenes, the system evaluates visible candidates during target selection and then focuses on the tracked target for behavior control.

\begin{table}[t]
\centering
\small
\caption{\textbf{Real-time deployment latency on the robot-mounted NVIDIA Jetson Thor.} YOLO11n and RTMW-l are converted to TensorRT FP16 engines. The deployed pipeline operates at the 30\,FPS camera rate; reported latencies reflect per-frame compute time.}
\label{tab:latency}
\begin{tabular}{l c c}
\toprule
Module & Latency & Deployment \\
\midrule
YOLO11n person detection & 5\,ms & TensorRT FP16 \\
BoT-SORT tracking & 1\,ms & CPU/GPU tracking \\
RTMW-l pose estimation & 8\,ms & TensorRT FP16 \\
Skeleton gesture model & 11\,ms & PyTorch \\
\midrule
Total, 1 person & $\mathbf{25}$\,ms & real-time @ 30\,FPS \\
Total, 3 persons & $\mathbf{42}$\,ms & real-time @ 24\,FPS \\
\bottomrule
\end{tabular}
\end{table}

For deployment adaptation, interaction segments flagged by the uncertainty score $u$ in Eq.~\eqref{eq:uncertainty} are merged and segmented by a cloud video understanding model (Gemini 2.5 Flash~\cite{gemini}) into 271 single-action clips using timestamp boundaries. Clips shorter than 3 seconds are discarded, since short fragments produced by window merging lack sufficient temporal context, leaving 71 clips. Against human verification, the cloud labels are correct on 67 of these ($94.4\%$); the 4 mislabeled clips are removed before fine-tuning, consistent with our practice of validating pseudo-labels before any model update. The classifier is then expanded from 7 to 10 classes and fine-tuned on a mixture of original human-labeled skeletons and the 67 pseudo-labeled deployment clips.

\begin{figure}[t]
    \centering
    \incfig{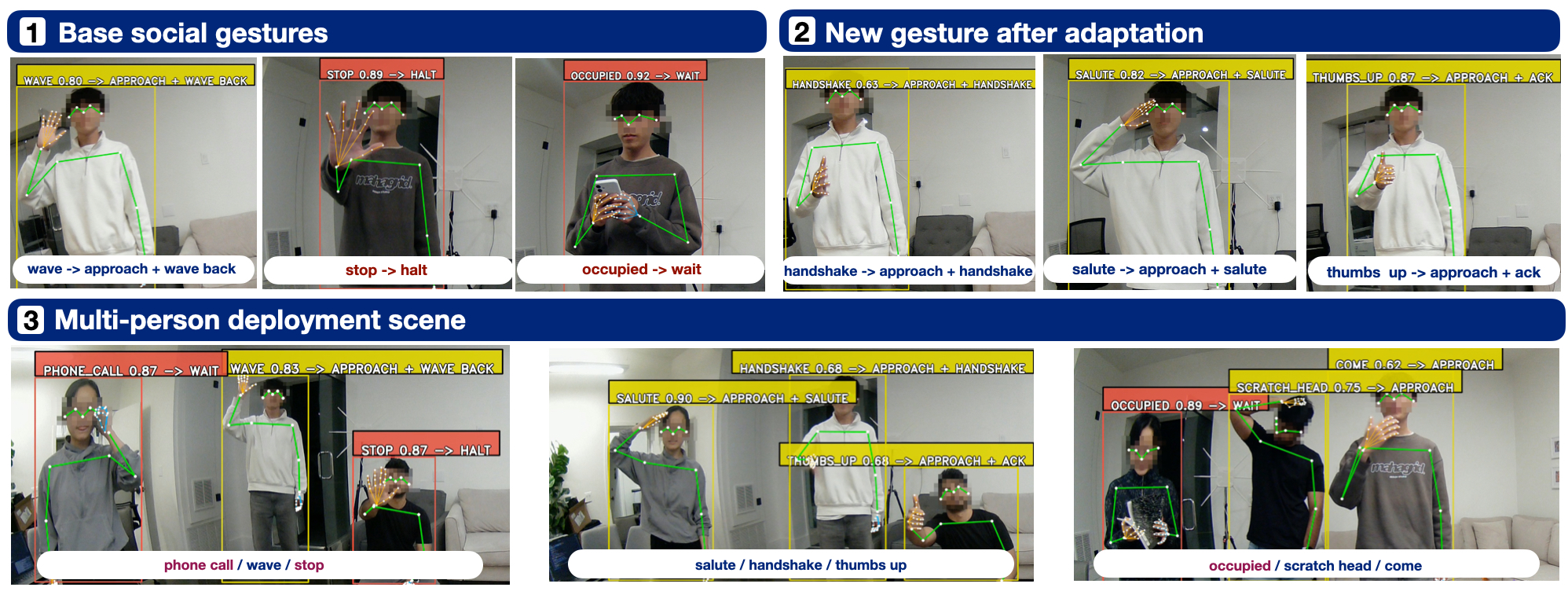}{62mm}
    \caption{\textbf{Robot-view recognition examples.} The examples include original social gestures, newly added gestures recognized after offline adaptation, and multi-person deployment scenes. Faces are blurred for privacy; the deployed system operates on the original stream.}
    \label{fig:qualitative_results}
\end{figure}

Table~\ref{tab:adaptation} shows that adaptation improves overall 10-class clip accuracy from $64.6\%$ to $87.9\%$ and macro F1 from $56.7\%$ to $88.5\%$. Accuracy on the original 7 classes remains unchanged at $98.5\%$, indicating that rehearsal preserves the initial gesture vocabulary, while the adapted model reaches $67.6\%$ accuracy on the three newly added classes. This is important for deployment because adding gestures should not weaken safety-critical behaviors such as \textit{stop} and \textit{wait}; the remaining new-class gap mainly reflects limited pseudo-labeled data and hand-localization difficulty. Figure~\ref{fig:qualitative_results} shows qualitative robot-view examples, including newly added gestures and multi-person scenes.

\begin{table}[t]
\centering
\small
\caption{\textbf{Offline deployment adaptation on a held-out 5-subject 10-class test set.} The base model recognizes only the original 7 classes and is evaluated in the expanded 10-class label space; new-class accuracy is therefore unsupported before adaptation. Fine-tuning expands the classifier using pseudo-labeled deployment skeletons.}
\label{tab:adaptation}
\begin{tabular}{l c c c}
\toprule
Metric & Base & Fine-tuned & $\Delta$ \\
\midrule
Overall 10-class clip accuracy & 64.6 & $\mathbf{87.9}$ & $\mathbf{+23.3}$ \\
Clip macro F1 & 56.7 & $\mathbf{88.5}$ & $\mathbf{+31.8}$ \\
Original 7-class clip accuracy & 98.5 & $\mathbf{98.5}$ & +0.0 \\
New 3-class clip accuracy & -- & $\mathbf{67.6}$ & -- \\
\bottomrule
\end{tabular}
\end{table}

\subsection{Efficiency Against Skeleton Recognition Baselines}
Table~\ref{tab:efficiency} compares SocioGesture with representative skeleton action recognition methods on NTU RGB+D 60. While many high-performing methods use multi-stream ensembles, SocioGesture uses a dual-branch body-hand recognizer evaluated in a single forward pass, with 1.3M parameters and 2.3 GFLOPs, operating in a practical efficiency regime for robot deployment. This comparison is intended to contextualize efficiency rather than claim state-of-the-art NTU accuracy.

\begin{table}[t]
\centering
\small
\setlength{\tabcolsep}{6pt}
\caption{\textbf{Computational efficiency on NTU RGB+D 60 under the cross-subject protocol.} Baseline values are reported from the corresponding papers. ``Passes'' denotes the number of inference passes or ensemble streams used for prediction. SocioGesture has body and hand branches internally, but is evaluated as a single forward pass for low-latency robot deployment.}
\label{tab:efficiency}
\begin{tabular}{l c c c c}
\toprule
Method & Passes & Params (M) & GFLOPs & X-Sub Acc. \\
\midrule
CTR-GCN~\cite{ctrgcn}        & 4 & 5.8  & 7.9   & 92.4\% \\
InfoGCN~\cite{infogcn}       & 6 & 9.4  & 10.0  & 93.0\% \\
BlockGCN~\cite{blockgcn}     & 4 & 5.2  & 6.5   & 93.1\% \\
PoseConv3D~\cite{poseconv3d} & 1 & 4.0  & 31.8  & 94.1\% \\
BHaRNet-P~\cite{bharnet}     & 2 & 9.06 & 3.42  & 96.3\% \\
\midrule
SocioGesture (ours)          & \textbf{1} & \textbf{1.3} & \textbf{2.3} & 94.6\% \\
\bottomrule
\end{tabular}
\end{table}

\subsection{Live Closed-Loop Deployment}
\label{sec:live}
We evaluate the complete closed-loop system on a Unitree G1 with five unseen participants who are not part of training, fine-tuning, or offline test data. In a greeter-robot protocol, each participant performs each of the 10 gestures three times with a fixed confidence threshold of $0.7$, producing 150 trials.

Table~\ref{tab:live_deployment} reports the results. The system recognizes the performed gesture correctly in 146/150 trials ($97.3\%$), and the adapted gestures transfer to live interaction nearly as reliably as the original gestures (44/45 vs.\ 102/105). Live recognition exceeds the offline clip-level number (Table~\ref{tab:adaptation}) because live trials are performed as discrete, well-segmented single gestures, whereas the offline test includes harder continuous clips. Behavior success is lower ($79.3\%$ overall) because the controller intentionally withholds action when confidence is below threshold; among trials in which the robot acts, the executed behavior is correct in $98.3\%$ of cases, and no \textit{stop} gesture causes the robot to approach. The dominant failure mode is therefore conservative non-action rather than incorrect action, which is the preferable trade-off for a greeter robot in early deployment.

\begin{table}[t]
\centering
\small
\setlength{\tabcolsep}{6pt}
\caption{\textbf{Live closed-loop deployment on the Unitree G1} with five unseen participants, 150 trials, fixed confidence threshold $0.7$. Recognition reports correctly recognized gestures regardless of confidence; behavior success reports trials in which the robot executed the correct behavior.}
\label{tab:live_deployment}
\begin{tabular}{l l c c}
\toprule
Gesture & Robot behavior & Recognition & Behavior success \\
\midrule
\textit{come} & Approach & 15/15 & 13/15 \\
\textit{wave} & Wave back & 15/15 & 14/15 \\
\textit{stop} & Halt & 15/15 & 14/15 \\
\textit{phone call} & Wait & 13/15 & 11/15 \\
\textit{occupied} & Wait & 15/15 & 15/15 \\
\textit{idle} (facing) & Approach (engage) & 14/15 & 13/15 \\
\textit{scratch head} (facing) & Approach (engage) & 15/15 & 10/15 \\
\textit{handshake} & Approach + handshake & 15/15 & 9/15 \\
\textit{salute} & Approach + salute & 15/15 & 14/15 \\
\textit{thumbs up} & Approach + acknowledgment & 14/15 & 6/15 \\
\midrule
Original 7 & -- & 102/105 & 85.7\% \\
New 3 & -- & 44/45 & 64.4\% \\
\textbf{Overall} & -- & \textbf{146/150} & \textbf{79.3\%} \\
\bottomrule
\end{tabular}
\end{table}

% =====================================================================
\section{Limitations and Future Work}
% =====================================================================
SocioGesture is designed around a deployment trade-off: the online system must remain lightweight and conservative for real-time robot behavior, while model improvement happens offline from uncertain interaction segments. This works well for short social gestures, but longer activities, multi-step exchanges, or context-dependent gestures may require temporal segmentation and memory beyond the current recognition window. Severe occlusion can also remove the hand or arm evidence needed to distinguish gestures; SocioGesture can defer action under low confidence, but cannot recover information absent from the input.

The adaptation loop reflects a safety--scalability trade-off. Offline video labeling and human verification reduce the risk of incorrect pseudo-label updates, but limit how quickly the system can learn from large amounts of interaction data. Future work should study scalable verification and continual adaptation, such as agreement across independent models, automatic label-consistency checks, and monitoring after redeployment. More broadly, gestures are short-horizon visual cues rather than complete social intent; future systems should combine gesture perception with speech, gaze, task state, and interaction history while preserving the lightweight onboard path.

% =====================================================================
\section{Conclusion}
% =====================================================================
We presented SocioGesture, a real-time and adaptive social gesture perception system for HRI. It combines confidence-aware body-hand skeleton encoding, lightweight temporal fusion, occlusion-aware training, conservative behavior selection, and uncertainty-driven offline adaptation. Experiments show strong held-out-subject recognition, robustness under structured occlusion, real-time edge deployment, vocabulary expansion from verified deployment clips, and successful live robot behavior. These results suggest a practical path toward social perception systems that are lightweight, robust to noisy deployment, and adaptive after deployment.

% =====================================================================
\bibliographystyle{unsrtnat}
\bibliography{references_checked}
% =====================================================================

\clearpage
\appendix

% =====================================================================
\section{Additional NTU Occlusion Stress Test}
\label{app:additional}
% =====================================================================
To separate HRI-specific effects from general skeleton robustness, we evaluate on NTU RGB+D 60~\cite{ntu_rgbd} under the cross-subject protocol. The NTU experiments use 2D skeletons extracted by RTMW~\cite{rtmw}, rather than the dataset's standard 3D Kinect skeletons, and apply controlled joint corruption to full-body 59-joint skeletons. These ablations use a simplified single-stream ST-GCN++ setting to isolate the effects of confidence-aware input channels, occlusion-aware training, and a confidence-gated graph variant (CWGCN). They are therefore intended as a robustness stress test rather than as the primary HRI result or an identical-input comparison to the main dual-branch SocioGesture recognizer. Clean accuracy is essentially unchanged under occlusion-aware training ($94.4\%\rightarrow94.5\%$), while occlusion robustness improves substantially.

\begin{table}[h]
\centering
\small
\setlength{\tabcolsep}{3.2pt}
\caption{\textbf{Occlusion robustness ablation on NTU RGB+D 60} using RTMW-extracted 2D 59-joint skeletons under the cross-subject protocol. The 9ch input uses the full confidence-aware representation in Eq.~\eqref{eq:skeleton_feature}; the 8ch input removes the confidence channel. Structured occlusions approximate common robot-mounted pose failures, while random joint masking is included as a controlled stress test. Avg. Occl. is the mean of the six occlusion columns.}
\label{tab:ntu_occlusion_ablation}
\resizebox{\linewidth}{!}{
\begin{tabular}{l l l c ccc ccc c}
\toprule
\multirow{2}{*}{Input} & \multirow{2}{*}{Backbone} & \multirow{2}{*}{Training} & \multirow{2}{*}{Clean} & \multicolumn{3}{c}{Structured occlusion} & \multicolumn{3}{c}{Random stress test} & \multirow{2}{*}{Avg. Occl.} \\
\cmidrule(lr){5-7}\cmidrule(lr){8-10}
& & & & Hand & Arm & Temp-Hand & R20 & R40 & R60 & \\
\midrule
8ch & ST-GCN++ & Standard & 94.3 & 79.5 & 49.3 & 91.0 & 19.6 & 4.5 & 2.9 & 41.1 \\
8ch & ST-GCN++ & Occlusion Aug. & 94.3 & 88.1 & 82.6 & 92.3 & 90.5 & 79.8 & 49.1 & 80.4 \\
8ch & ST-GCN++ & Occlusion Consistency & 94.4 & 88.2 & 83.5 & 92.7 & 91.6 & 83.3 & 55.5 & 82.5 \\
\midrule
9ch & ST-GCN++ & Standard & 94.4 & 81.8 & 27.9 & 91.5 & 12.3 & 3.4 & 2.5 & 36.6 \\
9ch & ST-GCN++ & Occlusion Aug. & 94.5 & 83.3 & 81.9 & 92.1 & 91.3 & 80.4 & 48.2 & 79.5 \\
9ch & ST-GCN++ & Occlusion Consistency & 94.5 & 87.8 & 82.8 & 92.7 & 92.0 & 84.6 & \textbf{61.7} & \textbf{83.6} \\
\midrule
9ch & CWGCN & Standard & 94.6 & 86.2 & 31.1 & 92.3 & 14.6 & 3.8 & 2.7 & 38.5 \\
9ch & CWGCN & Occlusion Consistency & \textbf{94.6} & 87.8 & \textbf{83.7} & 92.5 & 91.4 & 81.8 & 54.5 & 82.0 \\
\bottomrule
\end{tabular}}

\vspace{4pt}
{\footnotesize R20/R40/R60 denote random masking of 20\%, 40\%, and 60\% of joints. Hand masks both hands, Arm masks one arm and its connected hand, and Temp-Hand masks hand joints over a contiguous temporal span.}
\end{table}

As shown in Table~\ref{tab:ntu_occlusion_ablation}, standard models maintain high clean accuracy but collapse under arm occlusion and strong random joint corruption (e.g., 9-channel arm accuracy drops to 27.9\%, and R60 to 2.5\%). Occlusion augmentation recovers most of this robustness, and occlusion consistency further improves the most severe random-corruption cases. The confidence-gated graph variant (CWGCN) offers an interpretable reliability-aware mechanism, but the largest gains come from training with corrupted skeleton views rather than from the architecture itself.

% =====================================================================
\section{Implementation Details}
\label{app:implementation}
% =====================================================================
Table~\ref{tab:hyperparams} summarizes the hyperparameters used for training, deployment, and adaptation. All values are fixed before evaluation.

\begin{table}[h]
\centering
\small
\setlength{\tabcolsep}{5pt}
\caption{\textbf{Hyperparameters used for SocioGesture training, deployment, and adaptation.} Symbols match the Method section.}
\label{tab:hyperparams}
\begin{tabularx}{\textwidth}{@{}p{0.32\textwidth} c X@{}}
\toprule
Setting & Value & Notes \\
\midrule
\multicolumn{3}{@{}l}{\textit{Architecture}} \\
$d$ (feature width) & 256 & Per-stream feature dimension \\
Cross-attention heads & 4 & Multi-head temporal attention \\
$J_B/J_H$ (HRI) & 11 / 42 & Upper-body / hand joints \\
$J_B/J_H$ (NTU) & 17 / 42 & Full-body / hand joints \\
Face feature dim. & 3 & Yaw, pitch, confidence \\
\midrule
\multicolumn{3}{@{}l}{\textit{Training}} \\
Window size / stride & 16 / 8 & 50\% overlap \\
Optimizer & SGD & Momentum 0.9, weight decay $10^{-4}$ \\
Learning rate & 0.05 & Cosine schedule with 5-epoch warmup \\
Batch size & 64 & -- \\
Epochs & 80 & -- \\
Seed & 42 & Fixed for all reported runs \\
$\lambda_{\mathrm{aux}}$ & 0.3 & Auxiliary-head weight in Eq.~\eqref{eq:supervised_loss} \\
$\alpha_{\mathrm{occ}}$ & 0.5 & Occluded-view weight in Eq.~\eqref{eq:full_loss} \\
$\alpha_{\mathrm{cons}}$ & 0.5 & Consistency weight in Eq.~\eqref{eq:full_loss} \\
$\tau$ & 2.0 & Temperature in Eq.~\eqref{eq:consistency} \\
$E_w$ & 20 & Consistency warmup epochs \\
\midrule
\multicolumn{3}{@{}l}{\textit{Deployment uncertainty score}} \\
$\alpha$ & 0.5 & Entropy term $H(\mathbf{p})/\log K$ \\
$\beta$ & 0.3 & Body--hand disagreement term \\
$\gamma$ & 0.2 & Confidence term $1-\bar{c}$ \\
Mining threshold & 0.3 & Save segments with $u>0.3$ for offline labeling \\
Behavior threshold & 0.7 & Fixed confidence threshold for robot action \\
\midrule
\multicolumn{3}{@{}l}{\textit{Live deployment}} \\
Action cooldown & 5\,s & Minimum interval between robot actions \\
Approach distance & 1.0\,m & Social distance limit \\
Face-presence threshold & 0.6 / 0.4 & Enter / hold thresholds with EMA hysteresis \\
\midrule
\multicolumn{3}{@{}l}{\textit{Adaptation}} \\
Fine-tune learning rate & 0.005 & -- \\
Fine-tune epochs & 50 & -- \\
Backbone freeze & 5 epochs & Initial freeze before full fine-tuning \\
Pseudo-label clip min. & 3\,s & Shorter clips are discarded \\
\bottomrule
\end{tabularx}
\end{table}

\end{document}